%% file: root.tex
\pdfoutput=1

\documentclass[dvipsnames]{article}
\usepackage{arxiv}

\usepackage{amsmath,amssymb,amsthm,amsfonts,mathtools}
\usepackage{graphicx}
\usepackage{textcomp}
\usepackage{braket}
\usepackage{booktabs}
\usepackage{multirow}
\usepackage{xspace}

\usepackage{bm}

\usepackage{csquotes}

\DeclareMathAlphabet{\mathdutchcal}{U}{dutchcal}{m}{n}
\SetMathAlphabet{\mathdutchcal}{bold}{U}{dutchcal}{b}{n}
\DeclareMathAlphabet{\mathdutchbcal}{U}{dutchcal}{b}{n}
\DeclareMathAlphabet{\mathpzc}{OT1}{pzc}{m}{it}

\makeatletter
\let\MYcaption\@makecaption
\makeatother
\usepackage{subcaption} %
\makeatletter
\let\@makecaption\MYcaption
\makeatother

\usepackage[
    backend=biber
  ,url=false, isbn=false, doi=false %
  , maxnames=4
  , maxbibnames=99
  , minnames=3
  ,date=year                %
  ,labeldate=year
  , dashed=false
  , uniquename=false
  , style=authoryear
  ]{biblatex}
\renewrobustcmd*{\bibinitdelim}{\,} %
\usepackage{lipsum}

\usepackage[boxed,ruled,vlined,linesnumbered]{algorithm2e}
\DontPrintSemicolon
\SetKwProg{Fn}{function}{}{}
\SetKwFunction{FnSampleFree}{SampleFree}
\SetKwFunction{FnRestartArm}{RestartArm}
\SetKwFunction{FnPickArm}{PickArm}
\SetKwFunction{FnRewire}{Rewire}
\SetKwFunction{FnNearest}{Nearest}
\SetKwFunction{FnRestartArm}{RestartArm}
\SetKwComment{Comment}{$\triangleright$\ }{}
\SetKwInput{KwInit}{Initialise}

\usepackage{cleveref}                                        %
\crefname{assumption}{assumption}{assumptions}
\crefname{problem}{problem}{problems}
\crefname{algorithm}{Alg.}{Algs.}
\Crefname{algorithm}{Algorithm}{Algorithms}
\crefname{figure}{Fig.}{Figs.} %
\crefformat{equation}{(#2#1#3)}
\crefrangeformat{equation}{(#3#1#4) to~(#5#2#6)}
\crefmultiformat{equation}{(#2#1#3)}%
{ and~(#2#1#3)}{, (#2#1#3)}{ and~(#2#1#3)}

\usepackage{microtype}

\microtypesetup{activate={true,nocompatibility},final,tracking=true,kerning=true,factor=1100,stretch=10,shrink=10}
\usepackage{etoolbox}
\makeatletter
\pretocmd{\NAT@citexnum}{\@ifnum{\NAT@ctype>\z@}{\let\NAT@hyper@\relax}{}}{}{}
\makeatother

\begin{document}

\newcommand{\shortheadtitle}{
  Psy-LLM: Large Language Models for Mental Health Psychological Services
  }

\title{
  Psy-LLM: Scaling up Global Mental Health Psychological Services with \\ AI-based Large Language Models
  }

\author{
    Tin Lai\thanks{%
    Correspondence: \texttt{tin.lai@sydney.edu.au}
}%
\And
    Yukun Shi%
\And
    Zicong Du%
\And
    Jiajie Wu%
\And
    Ken Fu%
\And
    Yichao Dou%
\And
    Ziqi Wang%
\AND%
\normalfont
  School of Computer Science\\
  The University of Sydney\\
  Australia
}

\maketitle

\begin{abstract}
  The demand for psychological counselling has grown significantly in recent years, particularly with the global outbreak of COVID-19, which has heightened the need for timely and professional mental health support.
  Online psychological counselling has emerged as the predominant mode of providing services in response to this demand.
  In this study, we propose the Psy-LLM framework, an AI-based assistive tool leveraging Large Language Models (LLMs) for question-answering in psychological consultation settings to ease the demand for mental health professions.
  Our framework combines pre-trained LLMs with real-world professional Q\&A from psychologists and extensively crawled psychological articles.
  The Psy-LLM framework serves as a front-end tool for healthcare professionals, allowing them to provide immediate responses and mindfulness activities to alleviate patient stress.
  Additionally, it functions as a screening tool to identify urgent cases requiring further assistance.
  We evaluated the framework using intrinsic metrics, such as perplexity, and extrinsic evaluation metrics, with human participant assessments of response helpfulness, fluency, relevance, and logic.
  The results demonstrate the effectiveness of the Psy-LLM framework in generating coherent and relevant answers to psychological questions.
  This article discusses the potential and limitations of using large language models to enhance mental health support through AI technologies.
\end{abstract}

\newcommand{\citep}{\autocite}

\input{src/1.introduction}

\input{src/2.literature}

\input{src/3.research}

\input{src/4.methodologies}
\input{src/7.results}

\input{src/8.discussion}

\input{src/9.limitations}

\printbibliography

\end{document}

%% file: src/1.introduction.tex
\section{Introduction}

The field of AI utilising dialogue technology has witnessed significant growth, particularly in the domain of automatic chatbots and ticket support systems~\citep{handoyo2018ticketing}.
This application of dialogue technology has emerged as a cutting-edge and increasingly popular approach in the realm of AI-powered support systems.
With changing global dynamics, the severity of the ongoing pandemic, and an upsurge in psychological challenges the public faces, the mental well-being of young individuals, in particular, is a cause for concern.
The pressures of urbanisation and the internet have led to various psychological issues~\citep{trivedi2008rapid}, including depression, procrastination, anxiety, obsessive-compulsive disorder, and social phobia~\citep{tian2020psychological}, which have become prevalent ailments of our time.

Psychological counselling involves the utilisation of psychological methods to provide assistance to individuals experiencing difficulties in psychological adaptation and seeking solutions.
The demand for psychological counselling has witnessed a significant surge in recent years~\citep{chen2022mental}, while the availability of professional psychological consultants remains insufficient.
The profession of psychological consulting imposes high standards and qualifications.
For instance, registered psychologists within Psychological Associations require students to possess a master's degree in psychology-related disciplines, undergo a minimum of 150 hours of direct counselling, and receive face-to-face supervision by registered supervisors for no less than 100 hours~\citep{gay2021school}.
Additionally, the burnout rate among mental health professionals further exacerbates this shortage~\citep{joshi2020burnout}.

In 2020, the global outbreak of COVID-19 exacerbated the need for timely and professional psychological counselling due to the tremendous stress it imposed on society~\citep{kontoangelos2020mental}.
Consequently, online psychological counselling through the internet has progressively become the dominant mode of delivering counselling services~\citep{yurayat2023university}.
AI-based assistive psychological support not only addresses the severe supply-demand gap in the consulting industry but also enhances the responsiveness of online psychological counselling services, thereby promoting the implementation of mental health strategies.
Such an assistive tool serves to ease the shortage of mental health support when no human counsellors are available to help.

In light of these circumstances, our team is determined to develop an assistive mental health consulting framework to serve as a constant source of support.
Creating an AI-powered framework can allow users to engage with it comfortably, given its non-human identity, thereby reducing feelings of shame among users~\citep{Prochaska_2021}. In particular, with the absence of available psychological support, our framework serves as the second-best approach to providing timely support to patients.
Amid the challenges posed by the pandemic, online psychological counselling has proven instrumental and has gradually become the predominant form of counselling. However, the growing disparity between supply and demand within our society's psychological consultation industry is a pressing concern. The application of AI technology to mental health and psychological counselling is an emerging and promising field. Conversation frameworks, chatbots, and virtual agents are computer programs that simulate human conversation~\citep{deryugina2010chatterbots}. They can engage in natural and effective interactions with individuals, providing them with emotional experiences through the incorporation of emotional and human-like characteristics. In practical terms, dialogue frameworks hold significant potential for supporting the demand in online consultations and addressing supply-demand imbalances.

In this study, we propose an AI-based \textbf{Psy}chological Support with \textbf{L}arge \textbf{L}anguage \textbf{M}odels (\textbf{Psy-LLM}) framework designed for question-answering, with the purpose of providing online consultation services to alleviate the demand for mental health professionals during pandemics and beyond.
Psy-LLM is an online psychological consultation model pre-trained with Large Language Models (LLMs) and further trained with Q\&A from professional psychologists and large-scale crawled psychological articles.
The framework can give professional answers to users' requests for psychological support.
In particular, Psy-LLM can provide mental health advice both as recommendations for health professionals and as standalone tools for patients when no human counsellors are available due to time constraints or staff shortages.
Our model is built upon large-scale pre-training corpus models, specifically \emph{PanGu}~\citep{Zeng2021PanGuLA} and \emph{WenZhong}~\citep{fengshenbang}. The \emph{PanGu} model, developed by Huawei's Pengcheng Laboratory, and the \emph{WenZhong} model, developed by the Idea Research Institute, served as the basis for our work. For data acquisition, we collected a substantial number of Chinese psychological articles from public websites. Additionally, we obtained permission from the Artificial Intelligence Research Institute of Tsinghua University to utilise the PsyQA dataset, which comprises many question-answer pairs related to psychological counselling. Each answer in the dataset was reviewed and adjusted by professionals holding master's degrees or above in psychological counselling to ensure its quality. We fine-tuned the model in downstream tasks using the acquired dataset and PsyQA~\citep{psyqa}. As part of the evaluation process, we established a dedicated website and deployed the fine-tuned model on a server, allowing users to provide timely ratings. Based on the scoring results, we iteratively refined and re-fine-tuned the model.

Our contribution includes proposing a framework for AI-based psychological consultation framework and an empirical study on its effectiveness. We have successfully developed a mental health consulting model that effectively provides clear and professional responses to users' psychological inquiries. Empirically, we have tested deploying the model on a server, and the model can respond to users within seconds.
Our framework has the potential to offer a practical tool for professionals to efficiently screen and promptly respond to individuals in urgent need of mental support, thereby addressing and alleviating pressing demands within the healthcare industry.

%% file: src/2.literature.tex
\newcommand\citewithauthor[1] 
 {\citeauthor{#1}~(\citeyear{#1})~\citep{#1}}

\section{Related Works}

In recent years, there has been increasing interest in utilising AI for tackling difficult problems in traditional domains like 
adopting AI in the construction industry~\citep{regona2022opportunities},
localisation in robotic applications~\citep{lai2022slamreview},
assistance systems in the service sector~\citep{link2020use},
financial forecast~\citep{forexNonStationaryTimeSeries},
improving workflow in the oil and gas industry~\citep{koroteev2021artificial},
planning and scheduling~\citep{lai2022MEP},
monitoring ocean contamination~\citep{xu2022waterSedimentML},
remote sensing for search and rescue~\citep{lai2023UAV},
and even used in the life cycle of material discovery~\citep{li2020ai}.
Health care industry has been adopting AI-based machine-learning techniques for classifying medical images~\citep{castiglioni2021ai},
guiding cancer diagnosis~\citep{chugh2021survey},
as screening tools for diabetes~\citep{sensorsMLforDiabetes},
and ultimately improve the clinical workflow in the practice of medicine~\citep{brattain2018machine}.

One area of research focuses on using conversational agents, also known as chatbots, for mental health support. Chatbots have the potential to provide accessible and cost-effective assistance to individuals in need. 
For example, \citewithauthor{martinengo2022evaluation} qualitatively analysed user-conversational agents and found that these types of chatbots can offer anonymous, empathetic, and non-judgemental interactions that align with face-to-face psychotherapy. Chatbots can utilise NLP techniques to engage users in therapeutic conversations and provide personalised support. Results showed promising outcomes, indicating the potential effectiveness of chatbots in delivering mental health interventions~\citep{denecke2021artificial}.
Pre-trained language models have also gained attention in the field of mental health counselling. These models, such as GPT-3~\citep{brown2020language}, provide a foundation for generating human-like responses to user queries. \citewithauthor{wang2023prompt} explored the application of LLMs in providing mental health counselling. They found that LLMs demonstrated a certain level of understanding and empathy, providing responses that were perceived as helpful by users. However, limitations in controlling the model's output and ensuring ethical guidelines were highlighted.

Furthermore, there is a growing body of research on using NLP techniques to analyse mental health-related text data~\citep{gonzalez2017capturing}. Researchers have applied machine learning algorithms to detect mental health conditions~\citep{abd2020application}, predict suicidal ideation~\citep{ji2020suicidal}, and identify linguistic markers associated with psychological well-being~\citep{akstinaite2022identifying}. For instance, \citewithauthor{de2013predicting} analysed social media data to predict depression among individuals. By extracting linguistic features and using machine learning classifiers, they achieved promising results in identifying individuals at risk of depression.
Additionally, several studies have investigated the integration of modern technologies into existing mental health interventions. For instance, \citewithauthor{lui2017evidence} investigates the use of mobile applications to support the delivery of psychotherapy.

\citewithauthor{shaikh2022autonomous} developed a friendly AI-based chatbot using deep learning and artificial intelligence techniques. The chatbot aimed to help individuals with insomnia by addressing harmful feelings and increasing interactions with users as they experienced sadness and anxiety.
In another line of research, chatbots have been extensively studied in the domain of customer service. Many companies have adopted chatbots to assist customers in making purchases and understanding products. These chatbots provide prompt replies, enhancing customer satisfaction~\citep{9885724}.
Furthermore, advancements in language models such as BERT and GPT have influenced the development of conversational chatbots. Researchers have leveraged BERT-based question-answering models to improve the accuracy and efficiency of chatbot responses\citep{9652153}. The GPT models, including GPT-2 and GPT-3, have introduced innovations such as zero-shot and few-shot learning, significantly expanding their capabilities in generating human-like text\citep{Brown2020LanguageMA}. However, limitations in generating coherent and contextual responses and the interpretability of the models have been identified. The model incorporated a 48-layer Transformer stack and achieved a parameter count of 1.5 billion, resulting in enhanced generalisation abilities\citep{Brown2020LanguageMA}.

In summary, previous work in AI and NLP for mental health support has demonstrated the potential of chatbots, pre-trained language models, and data analysis techniques. These approaches offer new avenues for delivering accessible and personalised mental health interventions. Nonetheless, further research is needed to address ethical, privacy, and reliability issues and optimise integrating AI technologies into existing counselling practices.

%% file: src/3.research.tex
\section{Mental Health and Social Well-being in Overly Populated Cities}

The availability of mental health professionals has always been a major problem in overpopulated cities such as China.
The World Health Organisation has reported that the prevalence of depression in China exceeds 54 million people even before the onset of the COVID-19 pandemic~\citep{world2017depression}.
The situation has been exacerbated by the implementation of quarantine measures and social distancing, leading to a worsening condition~\citep{gou2022province}.
Unfortunately, only a small fraction of the affected population receives adequate medical treatment, as there are only 2 psychiatrists per 100,000 people in China~\citep{xiang2018rethinking}.
Consequently, there is a pressing need for a dynamic system that can assist patients effectively. Contemporary conversational chatbots have demonstrated their ability to emulate human-like conversations.

Hence, it is imperative to develop a user-friendly AI-based chatbot specifically designed to address anxiety and depression, with the aim of improving the user's emotional well-being by providing relevant and helpful responses.
This project aims to construct a Chinese psychological dialogue model capable of comprehending the semantic meaning of a consultant's request and offering appropriate advice, particularly to address the shortage of mental health workers during demanding periods. 
The trained model will be integrated into a website, featuring a user interface (UI) that ensures ease of operation, thereby enhancing the efficiency of psychological counselling.

\subsection{Research Questions}

Psychology is an intricate and advanced discipline gaining increasing significance as society progresses. However, due to its high barriers to entry, resources for psychological counselling have long been scarce. In numerous cases, individuals face challenges accessing adequate mental health support~\citep{19004}. Furthermore, the high cost of psychological counselling often prevents many individuals from prioritising their mental well-being. This issue is particularly prominent in China, a country with a large population where psychological problems have been historically overlooked. China needs a robust foundation for psychological counselling, including a deficient knowledge base and limited data. Consequently, intelligent assistance in psychology must be improved in the Chinese context.

Traditional psychological counselling primarily focuses on privacy and employs a one-on-one question-and-answer approach, inherently leading to inefficiencies. However, in today's high-pressure society, where mental health issues are pervasive, relying solely on scarce psychologists is arduous. Additionally, influenced by traditional culture, individuals often hesitate to acknowledge and address their psychological problems due to feelings of shame and perceiving such discussions as signs of weakness~\citep{sandhu}. This reluctance is especially prominent when conversing with real humans, let alone seeking assistance from unfamiliar psychologists.
Furthermore, with the advancement of modern Natural Language Processing (NLP) artificial intelligence models, there is a possibility to optimise the conventional and widely adopted question-and-answer model specifically for the field of psychology. When interacting with AI, people are more inclined to express their true thoughts and emotions without fear of prejudice and discrimination than real humans.
However, the lack of verbal cues and continuous monitoring of patients' emotional progress is also a cause of concern in an online psychological consultation context, even when performed by human counsellors~\citep {novella2022comparison}.

\subsection{Research Scope}
Through our project, we aim to make a meaningful contribution to the field of mental well-being. By leveraging our AI model and proposing a framework for an internet-accessible consultation, we intend to enhance the accessibility of mental health support, making it more affordable and providing an avenue for psychological question-and-answer interactions, particularly to address the shortage of human counsellors for mental health support.
To achieve this goal, we gather professional counselling question-and-answer data and psychologically relevant knowledge data to construct a robust question-and-answer model specific to this domain. The success of our project relies on the utilisation of high-quality question-and-answer models. We plan to employ established Chinese pre-trained models with exceptional human-computer interaction and communication skills, characterised by fluent language, logical reasoning, and semantic understanding. However, these existing pre-trained models need more specialised psychological expertise and emotional understanding for counselling purposes. To address this limitation, we intend to integrate two models, namely the \emph{WenZhong} model and the \emph{PanGu} model, and evaluate their performance to determine the more suitable choice as our final model.

Subsequently, it is imperative to make our model accessible to a broader audience. Leveraging the internet provides the most effective means to accomplish this objective. Together with the guidance of professional mental health experts, our model can provide an additional venue for the general public to access mental health support by easing the stress and demand on mental health staff through an open and online platform.

%% file: src/4.methodologies.tex
\section{Psy-LLM Framework}

The Psy-LLM framework aims to be an assistive mental health tool to support the workflow of professional counsellors, particularly to support those who might be suffering from depression or anxiety.

\subsection{Target Audience and Model Usage}

The contributing factor of Psy-LLM---an AI-powered conversational model is two-fold.
(1) Firstly, Psy-LLM is trained with a corpus of mental health supportive Q\&A from mental health professionals, which enables Psy-LLM to be used as an assistive tool in an online consultation context. 
When users want to seek support from an online chat, Psy-LLM can provide suggestive answers to human counsellors to ease the staff's workload.
Such an approach eases the entry barrier for newly trained mental health staff to provide useful and supportive comments for those in need.
(2) Furthermore, in the absence of human counsellors (e.g. during off-hours or high-demand periods), Psy-LLM can also be a web frontend for users to interact with the system in an online consultation manner.
Providing timely support to help-seeking individuals is especially important among suicidal individuals~\citep{hom2015evaluating}.
Therefore, an AI-powered online consultation might be the next best venue to respond to the absence of human counsellors.

\subsection{Large-scale Pre-trained LLMs}

Our project involves leveraging two large-scale pre-training models, namely \emph{WenZhong} and \emph{PanGu} to develop the question-answering language model. The utilisation of pre-training models offers several advantages, including:
(1) Enhanced Language Representations: Pre-training on extensive unlabeled data enables the model to acquire more comprehensive language representations, which in turn can positively impact downstream tasks.
(2) Improved Initialisation Parameters: Pre-training provides a superior initialisation point for the model, facilitating better generalisation performance on the target task and expediting convergence during training.
(3) Effective Regularisation: Pre-training acts as an effective regularisation technique, mitigating the risk of overfitting when working with limited or small datasets. This is especially valuable as a randomly initialised deep model is susceptible to overfitting on such datasets.
By harnessing the advantages of pre-training models, we aim to enhance the performance and robustness of our question-answering language model for psychological counselling.

\subsection{PanGu Model}
\emph{PanGu} model is the first Chinese large-scale pre-training autoregressive language model with up to 200 billion~\citep{Zeng2021PanGuLA}. In an autoregressive model, the process of generating sentences can be likened to a Markov chain, where the prediction of a token is dependent on the preceding tokens. The \emph{PanGu} model, developed within the MindSpore framework, was trained using 2048 Ascend AI processors provided by Huawei and trained on a high-quality corpus of 1.1TB. It was officially released in April 2021 and has achieved the top rank in the Chinese Language Comprehension Benchmark (CLUE), a widely recognised benchmark for Chinese language comprehension~\citep{xu2020clue}.
\begin{figure}[tb]
    \centering
    \includegraphics[width=0.95\linewidth]{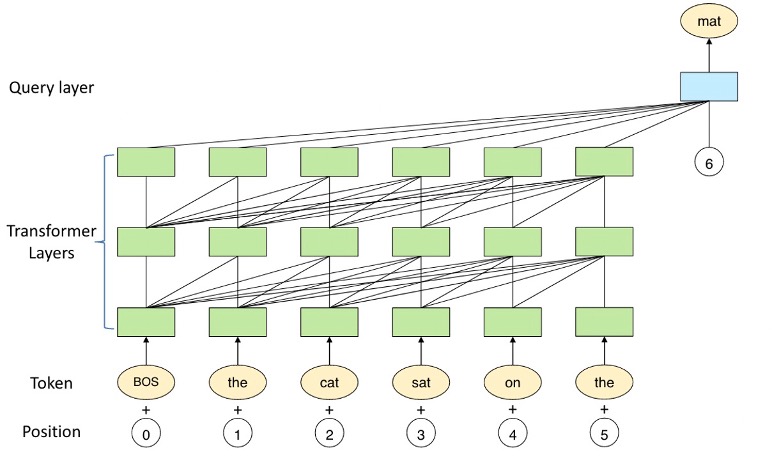}
    \caption{The model layers and architecture of the \emph{PanGu} model~\citep{Zeng2021PanGuLA}}
    \label{fig:PanGu}
\end{figure}

The architecture of the \emph{PanGu} model follows a similar structure to that of GPT-3, employing standard transformer layers~\cref{fig:PanGu}. Each transformer layer comprises two sub-layers: multi-head attention (MHA) and a fully connected feed-forward network (FFN). The MHA involves three primary steps: calculating the similarity between the Query and Key, applying a softmax function to obtain attention scores, and multiplying the attention scores with the Value to obtain the attention output. The attention output then passes through a linear layer and undergoes softmax to generate the output embedding. The output embedding is combined with the FFN input through a residual module. The FFN consists of two linear layers with a GeLU activation function between each consecutive layer. The MHA and FFN utilise the pre-layer normalisation scheme, facilitating faster and easier training of the Transformer model.

\begin{figure}[tb]
    \centering
    \includegraphics[width=0.6\linewidth]{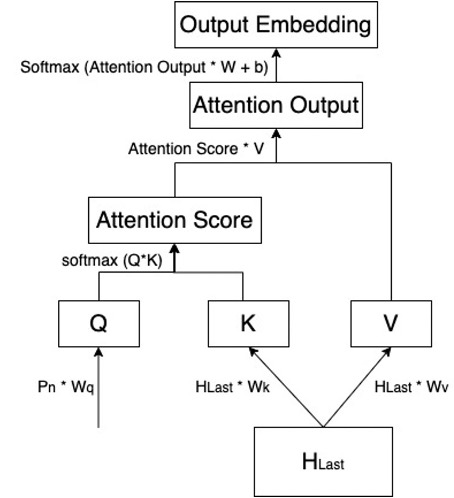}
    \caption{The Query Layer in the \emph{PanGu} Model}
    \label{fig:QueryArc}
\end{figure}

However, the last layer of the \emph{PanGu} model deviates from the standard transformer layer structure. Instead, it incorporates a query layer designed to predict the next token, thereby enhancing the model's positional awareness and improving generation effectiveness. The query layer is a narrow yet powerful decoder that relies solely on position information. The structure of the query layer is illustrated in~\cref{fig:QueryArc}. The primary distinction between the query layer and the transformer layer lies in the query input of self-attention. While the inputs of Query, Key, and Value in other self-attention layers of the transformer remain standard, the query layer introduces a query embedding, which functions similarly to position embedding, as the query input for self-attention in the last layer.

The \emph{PanGu} model is available in four distinct variations, each characterised by different parameter sizes (\cref{table:pangu-setting}). These variations include \emph{PanGu} 350M, \emph{PanGu} 2.6B, \emph{PanGu} 13B, and \emph{PanGu} 200B (which is not open source). The parameter sizes differ across these models, reflecting their varying levels of complexity and capacity for language understanding and generation.

\begin{table}[b]
    \centering
    \caption{The parametric size of the various settings in the \emph{PanGu} model. \label{table:pangu-setting}}
    \begin{tabular}{@{}cccccc@{}}
    \toprule
    Model & Parameters & Layers & Hidden Size & Head & Seq Length \\ \midrule
    \emph{PanGu} 350M & 350M  & 24 & 1024 & 16 & 1024\\
    \emph{PanGu} 2.6B  & 2.6B & 32 & 2560 & 40 & 1024   \\
    \emph{PanGu} 13B & 13.1B & 40 & 5120 & 40 & 1024 \\
    \emph{PanGu} 13B & 207.0B & 64 & 16384 & 128 & 1024 \\
    \bottomrule
    \end{tabular}
\end{table}

\subsection{WenZhong Model}
In addition to the \emph{PanGu} model, we have also incorporated the \emph{WenZhong} model as one of the models used. The \emph{WenZhong} model is a pre-trained model based on the GPT-2 architecture and trained on a large-scale Chinese corpus. Over the past few years, pre-trained models have become the foundation of cognitive intelligence, enabling advancements in natural language processing and computer vision algorithms.

The scale of pre-trained models has been rapidly increasing, growing by a factor of 10 each year, starting from the initial BERT model with 100 million parameters to the more recent GPT models with over 100 billion parameters. Given the nature of our task, which requires a generation model with expertise in different professional domains, we have opted for the \emph{WenZhong} model.

Due to the model size of LLMs like GPT, computing resources are the limiting factor hindering further progress in the field. Universities and research institutions often need more computing power to train and utilise large-scale pre-trained models. This limitation impedes the broader implementation of AI technologies. Hence, we have adopted the \emph{WenZhong} model, which is built upon a large pre-trained model trained on a Chinese corpus to avoid training the model from scratch.

The \emph{WenZhong} model series consists of one-way language models dominated by a Decoder structure and a series of powerful generation models. The \emph{WenZhong}-3.5B model, with 3.5 billion parameters, employs 100G data and 256 A100 GPUs for 28 hours of training, exhibiting strong generation capabilities. Thus, the \emph{WenZhong} model is highly powerful, featuring 30 decoder layers and billions of parameters. We have also utilised the \emph{WenZhong}-GPT2-110M version in this project, comprising 110 million parameters and 12 layers. It is important to note that the \emph{WenZhong} model has been pre-trained on the Wudao Corpus (300G version).

\subsection{Collecting Large Scale Dataset}
Two types of data sources were obtained for this project. The first dataset, PsyQA~\citep{psyqa}, consisting of question and answer pairs, focuses on Chinese psychological health support. The authors authorised us to use this dataset containing 22,000 questions and 56,000 well-structured, lengthy answers. The PsyQA dataset includes numerous high-quality questions and answers related to psychological support, and it had already undergone basic cleaning before we received it. We selected a test set of 5,000 samples from this PsyQA dataset for our experiments.

\subsubsection{Data Crawling}
The second dataset was obtained by crawling various Chinese social media platforms, such as Tianya, Zhihu, and Yixinli. These platforms allow users to post topics or questions about mental and emotional issues, whereas other users can also help to respond to help-seeking individuals. The Yixinli website specifically focuses on professional mental health support but only provides approximately 10,000 samples. Other types of datasets collected from these platforms included articles and conversations, which we converted into a question-and-answer format. However, we excluded the articles from our fine-tuning training due to the model's input limitations and the fact that our predictions focused on mental health support answers. The articles were often lengthy, and many of them were in PDF format, requiring additional time for conversion into a usable text format.
Consequently, we only obtained around 5,000 article samples. In order to address the lack of emotional expression in the text of these articles, we incorporated text data from oral expressions. We crawled audio and video data from platforms like Qingting FM and Ximalaya, popular audio and video-sharing forums in China. However, converting audio and video data into text format was time-consuming, resulting in a limited amount of data in our dataset. We utilised the dataset obtained from websites for fine-tuning training. Ultimately, our entire dataset consisted of 400,000 samples, each separated by a blank line, i.e., "\textbackslash n\textbackslash n".

\Cref{table:crawled-dataset} shows the time spent on data crawling from different websites. It is evident that most of the samples in this dataset were obtained from Tianya, resulting in a data size of approximately 2GB. The datasets from Zhihu and Yixinli were 500MB and 200MB, respectively. Overall, we spent approximately 70 hours on data collection. Although the data collected from the Internet was abundant and authentic, the cleaning process could have been smoother due to inconsistencies in the online data.

\begin{table}[tb]
    \centering
    \caption{dataset crawled from different platforms. \label{table:crawled-dataset}}
    \begin{tabular}{@{}ccc@{}}
    \toprule
    Platform & Data Size & Crawling Time  \\ \midrule
    Tianya & 2GB & 40h+  \\
    Zhihu & 500Mb & 20h+  \\
    Yixinli & 200Mb & 8h+ \\
    \bottomrule
    \end{tabular}
\end{table}

To address the time-consuming nature of web crawling, we implemented a distributed crawl technology that utilised idle computers connected to the Internet or other networks, effectively harnessing additional processing power~\citep{Thelwall2001AWC}. Our approach involved obtaining sub-websites from the main website and saving them using custom crawling code. This code primarily relied on Python libraries such as ``requests'', ``BeautifulSoup'', and ``webdriver''. In addition, we employed dynamic web crawlers capable of collecting clickable elements, simulating user actions, comparing web page states, manipulating the DOM tree, and handling various user-invoked events~\citep{Li2018AutomaticallyCD}. Unlike static page structures that cannot handle dynamic local refresh and asynchronous loading~\citep{Yao2012AnAF}, dynamic crawlers could extract data from behind search interfaces.

The process of the dynamic crawler involved leveraging web developer tools within the browser to obtain XHR (XMLHttpRequest) information, which included requests containing headers, previews, and responses. We acquired relevant files by systematically searching through these layers of data and capturing network packets. After obtaining the sub-websites using static and dynamic crawling methods, we distributed them across multiple idle computers. Each computer was assigned specific sub-websites, and we collected project-related data using a combination of static and dynamic crawling techniques. Ultimately, we utilised eight computers for the crawling process, which took approximately 70 hours.

\subsubsection{Data Cleaning}
In line with the \emph{PanGu} paper~\citep{Zeng2021PanGuLA}, we adopted the original data cleaning method utilised in the \emph{PanGu} model. Additionally, we incorporated some additional cleaning steps. The following are the cleaning steps we employed:
\begin{enumerate}
    \item \emph{Removal of duplicate samples}: We eliminated any duplicate samples in the dataset to ensure data uniqueness.
    \item \emph{Removal of samples containing advertised keywords}: We excluded samples that contained specific keywords associated with advertisements or promotional content.
    \item \emph{Deletion of data with less than 150 characters}: Samples with less than 150 characters were removed from the dataset, as they were deemed insufficient for effective model training.
    \item \emph{Removal of URLs}: Any URLs present in the samples were eliminated to maintain the focus on text content.
    \item \emph{Removal of user names and post time}: User names, such as "\texttt{@jack}", and post timestamps, were removed from the samples, as they were considered irrelevant to the text content.
    \item \emph{Removal of repeated punctuation}: Instances of repeated punctuation marks, such as "!!!" or ".....", were removed from the samples to ensure cleaner and more concise text.
    \item \emph{Conversion of traditional Chinese to simplified Chinese}: All traditional Chinese characters were converted to simplified Chinese characters to standardise the text.
\end{enumerate}

Following the data cleaning process, the dataset could be directly inputted into the \emph{PanGu} model. However, for training with the \emph{WenZhong} model, the samples needed further processing. Specifically, all punctuations were removed, and the samples were tokenised to ensure a consistent length of 1000 tokens for compatibility with the \emph{WenZhong} model.

\subsubsection{Data Analysis}
Data analysis plays a crucial role in understanding the fundamental characteristics of textual data. In the context of the Chinese language, the exploratory data analysis (EDA) methods may be less diverse than those used for English. In this study, we primarily employed two common methods: word frequency analysis and sentence length analysis, to gain insights into the dataset.

\begin{figure}[tb]
    \centering
    \includegraphics[width=0.55\linewidth]{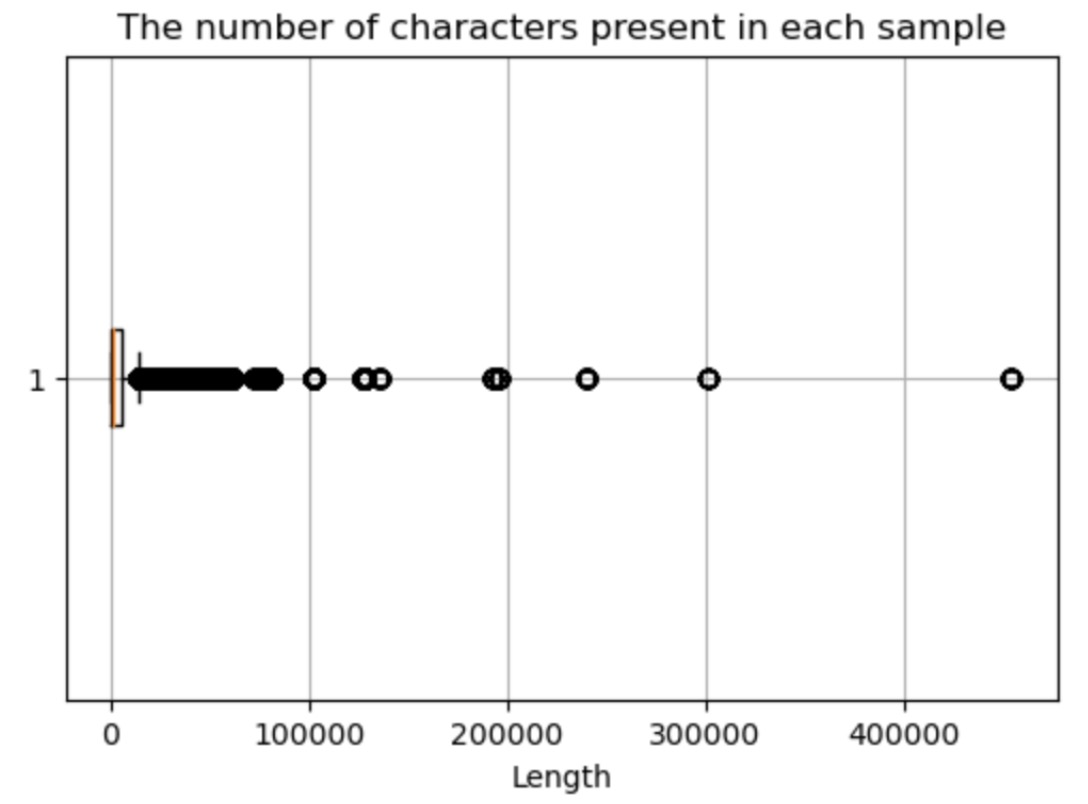}
    \caption{The number of characters in each sample}
    \label{fig:data-distribution}
\end{figure}

To analyse the distribution of characters in each sample, we referred to the character number data presented in~\cref{table:dataset-distribution}. By visualising this information using a box chart, we examined the range of character counts across the samples.
Some samples are empty after the data cleaning, which we then prune from our dataset.
\Cref{fig:data-distribution} displays the distribution of sample lengths, indicating that the majority of samples fall within the range of 10,000 characters.

Overall, these preliminary analyses allowed us to gain initial insights into the dataset and provided a foundation for further exploration and understanding of the textual data.

\begin{table}[tb]
    \centering
    \caption{Data distribution of the length of each sample. \label{table:dataset-distribution}}
    \begin{tabular}{@{}cccccccc@{}}
    \toprule
    Count & Mean & Std & Min & 25\% & 50\% & 70\% & Max \cr \midrule
    371,434 & 5,343 & 11,335 & 0 & 653 & 1,835 & 6,039 & 454,611 \cr
    \bottomrule
    \end{tabular}
\end{table}

\begin{figure}[tb]
    \centering
    \includegraphics[width=0.55\linewidth]{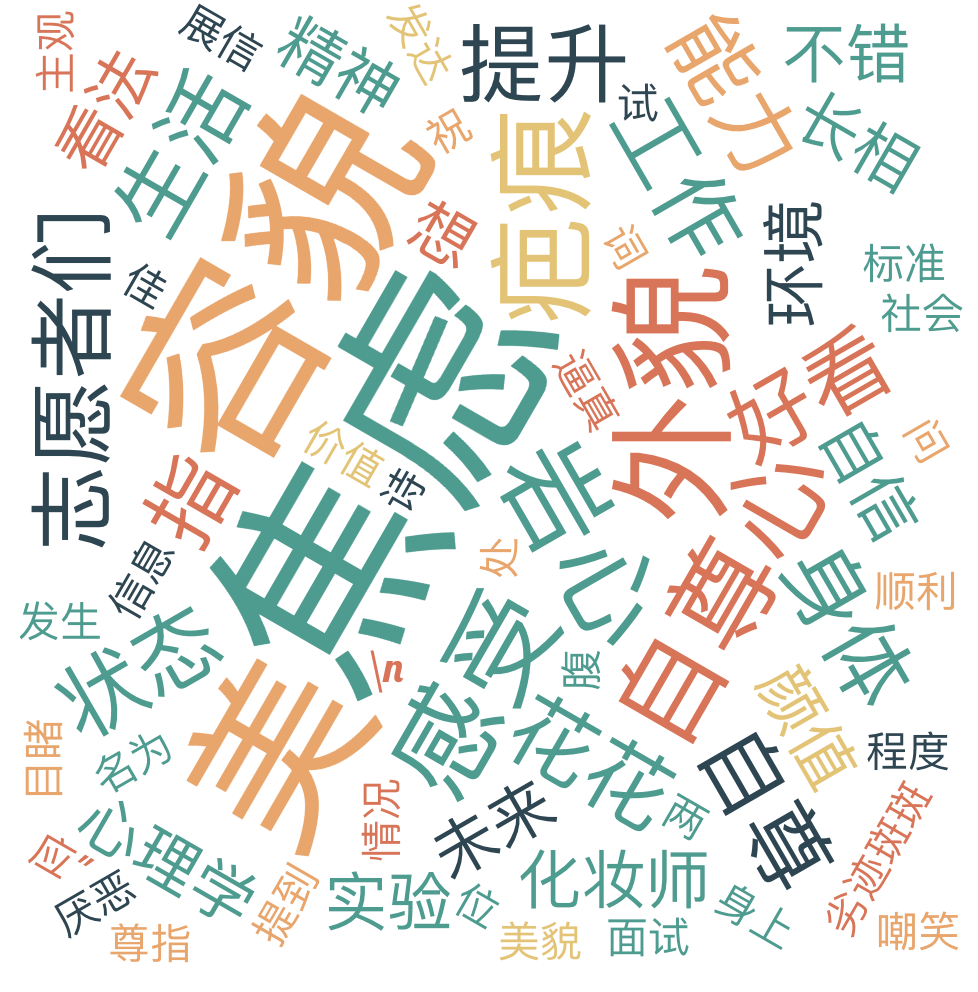}
    \caption{Word cloud of the frequent word within our dataset}
    \label{fig:frequent-word}
\end{figure}

To examine the word frequency in our dataset, we conducted an analysis after removing the stop words. The word cloud visualisation in~\cref{fig:frequent-word} illustrates the most frequent words in the dataset. Notably, the prominent words observed include ``Anxiety'', ``Appearance'', ``Marriage'', ``Relationship'', ``Family'', and ``Stressful''. These words are highly relevant to the topic of mental health, indicating that our dataset is robust and aligns well with the focus of our task.

The presence of these mental health-related terms further underscores the suitability of our data for addressing the objectives of our study. It suggests that our dataset encompasses significant content related to psychological aspects, allowing us to effectively explore and address relevant topics in the context of our research.

\subsection{Model Training}
\textbf{Model Size} We plan to use \emph{PanGu} 350M to generate language considering the computational power, which contains 350 million parameters, 24 layers, 1024 hidden sizes and 16 attention heads. Besides, we also want to train the \emph{WenZhong}-110M model that contains 12 layers and has 110M parameter.

\textbf{Training Data} We employ the 2.85GB psychology corpus data crawled from psychology platforms like Yixinli and Tianya, to train the original \emph{PanGu} 350M model. After that,  we use 5,6000 question-answer pairs from PsyQA dataset to fine-tune the model.

\textbf{Training Platform} We train the \emph{PanGu} model on the OpenI platform with a free V100 graphics card GPU because OpenI is the open source platform of \emph{PanGu} model, convenient for us to deploy the required files, image and GPU. The batch size is set to 8, and the training iteration is set to 100,000 because we found that 50,000 iterations is not enough for the model's loss to converge.
We train the \emph{WenZhong} model in jupyter notebook. To fine-tune this model, we need to tokenise the data, which can transform words into tokens. Besides, we also need to isolate the max length of each sentence as 500.

\subsection{Dataset Evaluation}

Determining the cleaning rules and data filtering thresholds are important aspects of the data cleaning process. We employed a data quality evaluation method that combined both manual and model-based evaluations to evaluate the dataset obtained from website crawling.

For the model-based evaluation, we utilised the \emph{PanGu} 350M model and calculated the perplexity metric after each data cleaning stage. A lower perplexity value indicates a more effective cleaning process and higher dataset quality.
In addition to the model-based evaluation, we sought input from experts in psychology. We invited two members from our University's School of Psychology, Faculty of Science, to perform a random sample check on the dataset after it had undergone the cleaning process. While this method does not cover the entire corpus comprehensively, it provides valuable insights and plays a role in data cleaning and quality evaluation.

The evaluation process involved the following steps: First, we provided the experts with a sample of the cleaned dataset and asked them to assess its quality based on their expertise and domain knowledge. They evaluated the dataset for accuracy, relevance, and coherence, providing feedback and suggestions for further improvements.

Next, we conducted a comparative analysis between the model-based evaluation and the expert evaluation. We examined the perplexity scores obtained from the \emph{PanGu} 350M model and compared them with the feedback provided by the experts. This allowed us to identify any discrepancies or areas of improvement in the dataset.

Overall, the combination of model-based evaluation and expert assessment comprehensively evaluated the dataset quality. It allowed us to identify and address any issues or shortcomings in the data cleaning process, ensuring that the final dataset used for training and evaluation was high quality and suitable for our research purposes.

\subsection{Model Training Setting}

\emph{Models:} For our training, we utilise the \emph{PanGu} 350M model, considering the available computational resources. This model consists of 350 million parameters, 24 layers, a hidden size 1024, and 16 attention heads. Additionally, we target the \emph{WenZhong}-110M model, which contains 12 layers and 110 million parameters.

\emph{Training Data:} We collected a psychology corpus dataset totalling 2.85GB, which was crawled from psychology platforms such as Yixinli and Tianya. This dataset was used for training the original \emph{PanGu} 350M model. Subsequently, we fine-tuned the model using 56,000 question-answer pairs from the PsyQA dataset.

\emph{Training Platform:} The \emph{PanGu} model was trained on the OpenI platform, utilising a free 1 V100 graphics card GPU. OpenI is an open-source platform specifically designed for the \emph{PanGu} model, allowing us to easily deploy the necessary files, images, and GPU resources.
For training with the V100 graphics card (32 GB memory), the minimum recommended configuration is one card, while the recommended configuration is two. The graphics card requirements can be adjusted based on the memory size (for example, a 16GB memory card would require twice as many cards as the V100). Increasing the number of graphics cards can help improve the training speed if the dataset is large.
We set the batch size to 8 and performed training for 100,000 iterations, as we observed that 50,000 iterations were insufficient for the model's loss to converge.
For the \emph{WenZhong} model, we used Jupyter Notebook to run the pre-trained model and fine-tuned it on a system with 64GB memory and an RTX3060 graphics card. The version details of the hardware and software components are listed in~\cref{table:hardware-software-version}.

\begin{table}[tb]
    \centering
    \caption{Hardware and Software Versions}
    \label{table:hardware-software-version}
    \begin{tabular}{@{}cc@{}}
    \toprule
    Hardware and Software & Version \\
    \midrule
    Operating System & Windows 10 \\
    numpy & 1.18.5 \\
    pandas & 1.3.4 \\
    torch & 1.11.0 \\
    tokenizers & 0.13.1 \\
    tqdm & 4.64.1 \\
    jupyter & 1.0.0 \\
    transformers & 4.23.1 \\
    \bottomrule
    \end{tabular}
\end{table}

\subsubsection{Training Process}
According to the guide of training \emph{PanGu} model with GPU, the first step is environment configuration. We prepared Pytorch, \emph{PanGu} image, one V100 graphics card and some \emph{PanGu} model files like vocabulary. The second step is data preprocessing. We put the training corpus into a text file, and each sample is a paragraph separated by two newlines, then converted it into a binary file because it is the required input format of the training \emph{PanGu} model. The third step is model training. We uploaded the \emph{PanGu} model and the bin file to the OpenI platform and set some parameters like iteration to train it. Our training procedure of \emph{PanGu} model consists of two steps. Firstly, we train the original \emph{PanGu} 350M model with all the crawling data for 100,000 iterations. The model starts to converge at about 60,000. This model has learned psychology domain knowledge based on pre-trained data. Secondly, we fine-tune it with the PsyQA dataset to improve the model's capability to provide useful answers for users on mental health support.

We used the early stop method to choose appropriate iterations. Stopping the training of the network before the validation loss increased effectively prevented the model from overfitting. For example, when the model has more than 9,000 training iterations, the validation loss of the model starts to rise, which means the phenomenon of overfitting occurs.
A similar approach is also used for the \emph{WenZhong} model.

\begin{figure}[tb]
    \centering
    \includegraphics[width=0.8\linewidth]{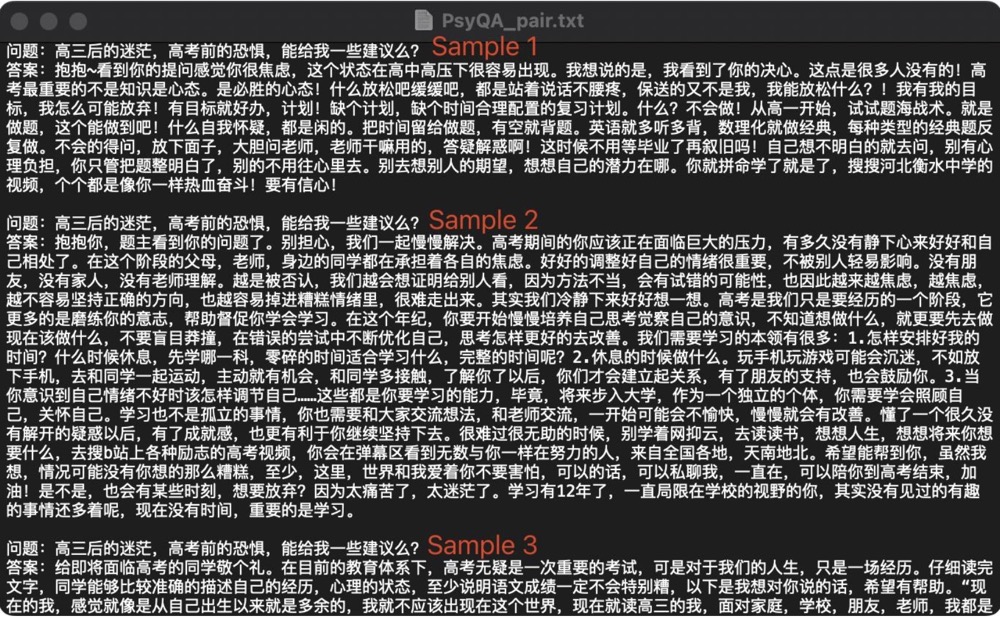}
    \caption{Prepareing the training corpus}
    \label{fig:mylabel}
\end{figure}

\subsection{Model Evaluation}

In this section, we assess the performance and effectiveness of our proposed language model for online psychological consultation. We employ a combination of intrinsic and human evaluation metrics to evaluate the model's capabilities comprehensively. We begin by utilising perplexity, ROUGE-L, and Distinct-n metrics to measure the model's language generation quality, similarity to the reference text, and diversity. Additionally, we recognise the limitations of these metrics and emphasise the importance of human evaluation in providing subjective assessments of the model's outputs, considering factors such as coherence, relevance, and overall quality. Through this comprehensive evaluation approach, we aim to gain a comprehensive understanding of our model's strengths, weaknesses, and suitability for its intended purpose in the context of online psychological consultation.

\subsubsection{Metric-based Evaluation}

Perplexity is a widely used intrinsic evaluation metric that measures how well a language model predicts a given sample. Mathematically, perplexity is defined as the reciprocal of the average probability assigned to each token in the dataset by the language model~\citep{Chen1998EvaluationMF}. In simpler terms, a lower perplexity value indicates better language model performance. Since perplexity is based on the average log-likelihood of the dataset, it can be computed quickly and is statistically robust, as it is not easily affected by outliers.

The formula for calculating perplexity is given by
 \begin{align}
    PP(W)
    &= \mathbb{P}\left( w_1 w_2 \ldots w_N \right)^{{-1}/{N}} \\
    &=\sqrt[N]{ \prod_{i=1}^N \frac{1}{\mathbb{P}(w_i)} }
 \end{align}
where $PP$ is the perplexity, $\mathbb{P}$ is the probability of the $i^{th}$ word, and $N$ is the length of a sentence.
It is important to note that perplexity tends to decrease as the dataset size increases, indicating better performance.

However, it is crucial to understand that low perplexity does not necessarily equate to high accuracy. Perplexity is primarily used as a preliminary measure and should not be solely relied upon for evaluating model accuracy. Additionally, comparing the performance of models on different datasets with varying word distributions can be challenging~\citep{Chen1998EvaluationMF}. Therefore, while perplexity provides valuable insights into model performance, it should be complemented with other evaluation metrics and considerations when assessing model accuracy.

ROUGE-L (Longest Common Subsequence) is an evaluation metric that measures the number of overlapping units between the predicted text generated by a language model and the actual reference text~\citep{Lin2004ROUGEAP}. ROUGE-L measures how closely the generated text matches the desired output by quantifying the similarity between the predicted and reference texts.

Distinct-1 and Distinct-2 are evaluation metrics that assess the diversity of the generated text. Distinct-1 calculates the number of distinct unigrams (individual words) divided by the total number of generated words. In contrast, Distinct-2 calculates the number of distinct bigrams (pairs of adjacent words) divided by the total number of generated bigrams~\citep{Li2016ADO}. These metrics reflect the degree of diversity in the generated text by quantifying the presence of unique unigrams and bigrams.

The formulas for calculating Distinct-n are as follows:

\begin{equation}
    \text{Distinct-n} := Distinct(n)=\frac{Count(\texttt{unique}, \texttt{n-gram})}{Count(\texttt{word})}
\end{equation}

Here, $Count(\texttt{unique}, \texttt{n-gram})$ represents the number of $n$-grams that are not repeated in a reply, and $Count(\texttt{word})$ indicates the total number of $n$-gram words in the reply. A higher value of $Distinct(n)$ indicates a greater diversity in the distinct generations.

These evaluation metrics, including perplexity, ROUGE-L, Distinct-1, and Distinct-2, provide insights into the quality, similarity, and diversity of the generated text by the language model. They serve as valuable tools for assessing the performance and effectiveness of the model in generating accurate and diverse outputs.

While perplexity and Distinct-n provide insights into the language model's performance in language generation, they do not necessarily indicate high accuracy. Therefore, in order to evaluate models more convincingly, human evaluation is still necessary. Human evaluators can provide subjective assessments of the generated text, considering factors such as coherence, relevance, and overall quality, which are important aspects that cannot be fully captured by automated evaluation metrics alone.

\subsubsection{Human evaluation}

For human evaluation, we have developed an online marking system to assess the performance of our language model in the context of online psychological consultation. This evaluation system aims to streamline the process and ensure effective assessment by focusing on four key metrics: Helpfulness, Fluency, Relevance, and Logic. Each metric is scored on a scale of 1 to 5, allowing evaluators to provide a quantitative assessment of each aspect.
The four metrics are defined as follows:

\begin{enumerate}
    \item \textbf{Helpfulness:} This metric evaluates whether the generated response is helpful for patients seeking psychological support.
    \item \textbf{Fluency:} Fluency refers to the degree of coherence and naturalness exhibited in the generated response.
    \item \textbf{Relevance:} Relevance assesses the extent to which the response's content directly relates to the posed question.
    \item \textbf{Logic:} Logic examines the logical consistency and coherence of the meaning conveyed in the generated response.
\end{enumerate}

To conduct the human evaluation, we invited six students from the psychological faculty to assess a set of 200 question-answer pairs generated by our model. We employed two evaluation methods to understand the model's performance comprehensively.

In the first method, evaluators compared responses generated by the \emph{PanGu} model and the \emph{WenZhong} model in response to the same question. They assigned scores to these answers based on the predetermined metrics, allowing for a direct comparison between the two models.
The second method involved incorporating the actual answers alongside the predicted responses as a whole, allowing evaluators to assess the differences and similarities between the generated and actual responses.

By employing these human evaluation methods, we aim to gain valuable insights into the performance of our language model, particularly in terms of the disparities between predicted and actual responses. This comprehensive evaluation approach will provide a deeper understanding of the model's capabilities and guide further improvements in its performance for online psychological consultation.

%% file: src/7.results.tex
\section{Experimental Results}

In this section, we present the findings and outcomes of the evaluation and experimentation conducted to assess the performance and effectiveness of our proposed language model for online psychological consultation. This section provides a comprehensive analysis of the model's performance based on intrinsic and human evaluation metrics. We discuss the results obtained from metrics such as perplexity, ROUGE-L, and Distinct-n, which shed light on language generation quality, similarity to reference text, and diversity of the generated responses. Additionally, we present the outcomes of the human evaluation, which includes scores given by evaluators based on metrics such as Helpfulness, Fluency, Relevance, and Logic. Through these rigorous evaluations, we aim to provide an in-depth understanding of the strengths and weaknesses of our language model and its suitability for the task of online psychological consultation.

\subsection{Result of Intrinsic Evaluation}

The results of the intrinsic evaluation comparing the performance of the \emph{PanGu} model and the \emph{WenZhong} model are presented in \cref{table:intrinsic-metric}. The metrics used for evaluation include perplexity, ROUGE-L, Distinct-1, and Distinct-2.

As shown in~\cref{table:intrinsic-metric}, the \emph{PanGu} model outperforms the \emph{WenZhong} model across all metrics. The \emph{PanGu} model achieves a lower perplexity value of 34.56 compared to 38.40 for the \emph{WenZhong} model, indicating that the \emph{PanGu} model better predicts the sample probabilities in the dataset.

Furthermore, the ROUGE-L score, which measures the similarity between the generated responses and the reference text, is higher for the \emph{PanGu} model (28.18) than the \emph{WenZhong} model (23.56). This suggests that the \emph{PanGu} model generates responses more aligned with the expected answers.

In terms of diversity in generated responses, the \emph{PanGu} model also exhibits higher Distinct-1 (4.57) and Distinct-2 (12.74) scores compared to the \emph{WenZhong} model (3.55 and 9.67, respectively). This indicates that the \emph{PanGu} model produces a wider range of distinct unigrams and bigrams in its responses, leading to a more diverse language generation.

The intrinsic evaluation results consistently demonstrate that the \emph{PanGu} model performs better than the \emph{WenZhong} model across all metrics. These findings highlight the superior language generation capabilities of the \emph{PanGu} model in the context of online psychological consultation.

\begin{table}[tb]
    \centering
    \caption{The result of Intrinsic Evaluation between two models. \label{table:intrinsic-metric}}
    \begin{tabular}{@{}ccccc@{}}
    \toprule
    Model & Perplexity & Rouge-L & Distinct1 & Distinct2  \\ \midrule
    WenZhong & 38.40  & 23.56 & 3.55 & 9.67\\
    PanGu & 34.56 & 28.18 & 4.57 & 12.74  \\
    \bottomrule
    \end{tabular}
\end{table}

\subsection{Case Study: Examining Psy-LLM's responses on coping with depression}

In the following, we will qualitatively evaluate the outputs of both models by asking our Psy-LLM a common question for individuals suffering from depression.

\subsubsection{PanGu Model}
\begin{figure}[tb]
    \centering
    \includegraphics[width=0.9\linewidth]{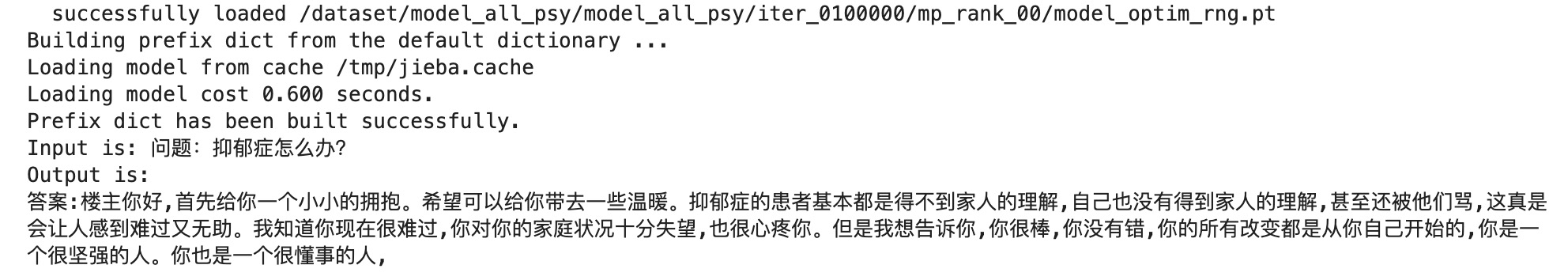}
    \caption{A demonstration of the \emph{PanGu} model's Q\&A capability}
    \label{fig:pangu-demo}
\end{figure}

To demonstrate the capabilities of the \emph{PanGu} model, we provide an example of question answering using the model on coping with depression (\cref{fig:pangu-demo}), which we have translated into English. The question posed to the model is,
\begin{displayquote}
\emph{``Question: How to deal with depression?''},
\end{displayquote}
and the model's response is as follows:
\begin{displayquote}
\emph{``Answer: Hello, first let me give you a small hug. I hope it brings you some warmth. Typically, individuals with depression are often misunderstood and even scolded by their families, which can be really disheartening. I understand that you are feeling very sad and disappointed with your family situation. I want you to know that you are a remarkable person, and you are not at fault. Remember, change starts with you, and you are incredibly strong and sensible.''}
\end{displayquote}

This response showcases the model's fluency and ability to employ different strategies.
Analysing the response in~\cref{fig:pangu-demo}, we can observe that the model utilises various strategies. The initial part, offering a virtual hug, demonstrates a reassurance strategy~\cite{psyqa}. It provides emotional support and encouragement to individuals seeking consultation, helping them relax. The second part, addressing family dynamics, employs a restatement strategy by clearly reiterating the question to provide a more realistic understanding of the individual's mental health condition. The final part utilises a guidance strategy, advising the individual on steps to improve their situation. Overall, the model's response is coherent and consistent with normal human logic, making it highly valuable.

The training of the \emph{PanGu} model consists of two steps. Initially, the model is trained with a vast amount of crawled data, enabling it to learn extensive domain knowledge in psychology. Subsequently, the model undergoes fine-tuning using the downstream PsyQA dataset. During this stage, the model primarily focuses on learning the pattern of \emph{``Question:''} and \emph{``Answer:''} to generate responses in the same format as the downstream tasks. When we input a question in the format \emph{``Question: $\cdots$''} to the model, the model outputs a response in the format \emph{``Answer: $\cdots$''}. This demonstrates that the model has successfully learned the desired pattern and can generate appropriate responses.

\subsubsection{WenZhong Model}

\begin{figure}[tb]
    \centering
    \includegraphics[width=0.9\linewidth]{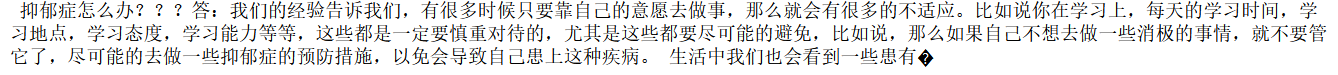}
    \caption{A demonstration of the \emph{WenZhong} model's Q\&A capability}
    \label{fig:WenZhong-demo}
\end{figure}

In \emph{WenZhong} model, after fine-tuning the model using Psychology data, the model can output very suitable answer.
We also utilised the \emph{WenZhong} model to answer the same question (\cref{fig:WenZhong-demo}), and the generated response was:

\begin{displayquote}
\emph{``Our experience tells us that there are many times when you just have to do things by your own will, then there will be a lot of discomfort. For example, if you are studying, the time of day, the place of study, the attitude of study, the ability to study, etc., these must be treated carefully, especially these should be avoided as much as possible, for example, then if you do not want to do something negative, leave it alone and do as much as possible to do some preventive measures for depression, so that it will not lead to the development of this disease.''}
\end{displayquote}

This example demonstrates that the \emph{WenZhong} model can provide suitable answers in the field of psychology. However, one challenge we encountered is that some generated outputs do not directly address the question. This issue may be attributed to the limited fine-tuning data specific to psychology. In order to further improve the performance of the \emph{WenZhong} model in psychology-related tasks, a larger and more diverse dataset from the field of psychology could be incorporated during the fine-tuning process.

\subsection{Human Evaluation}

To conduct an empirical evaluation of Psy-LLM's effectiveness, we enlisted the participation of six students from the psychological faculty to assess a set of 200 question-answer pairs generated by our language model. In order to obtain a comprehensive understanding of the model's performance, we employed two evaluation methods for the participants to provide ratings on the responses.
We have created a web front-end for users to access our Psy-LLM platform, and their technical details are discussed in~\cref{sec:web}.

The first method directly compared responses generated by both the \emph{PanGu} and the \emph{WenZhong} models in response to the same question. Evaluators assigned scores to these answers based on predetermined metrics, enabling a clear and direct comparison between the two models' performance.
In the second method, we presented evaluators with a combined set of predicted and actual responses. This allowed them to evaluate and assess the differences and similarities between the generated responses and the ground truth answers.

By utilising these human evaluation methods, we aim to gain valuable insights into the performance of our language model, particularly in terms of the disparities between predicted and actual responses. This comprehensive evaluation approach will provide a deeper understanding of the model's capabilities and guide further improvements in its performance for online psychological consultation.

The human evaluation results, using two different methods, are presented in~\cref{table:human-eval-1,table:human-eval-2}. These evaluate human-perceived metrics of \emph{Helpfulness, Fluency, Relevance, and Logic}.
\Cref{table:human-eval-1} shows the results of the first human evaluation method, where evaluators provided scores for each metric. Consistent with the findings from the intrinsic evaluation, the \emph{PanGu} model outperforms the \emph{WenZhong} model in terms of \emph{Helpfulness} (3.87 vs. 3.56), \emph{Fluency} (4.36 vs. 4.14), \emph{Relevance} (4.09 vs. 3.87), and \emph{Logic} (3.83 vs. 3.63). These results indicate that human evaluators generally consider the \emph{PanGu} model's generated responses more helpful, fluent, relevant, and logical than the \emph{WenZhong} model.

However, a notable observation is made when comparing the scores obtained in~\cref{table:human-eval-1} with the scores from~\cref{table:human-eval-2}. \Cref{table:human-eval-2} presents the scores for the predicted answers of both models as well as the actual answers. Interestingly, the scores for the actual answers are significantly higher than those for the predicted answers of both models across all metrics. This discrepancy suggests that the evaluators, who had the opportunity to compare the actual answers with the predicted answers, marked the predicted answers relatively lower. This finding highlights the importance of incorporating human evaluation in assessing the performance of language models and the need for further improvement in generating more accurate and satisfactory responses.

In summary, the human evaluation results align with the intrinsic evaluation findings, indicating that the \emph{PanGu} model performs better than the \emph{WenZhong} model. However, it is important to note that the scores for the actual answers are considerably higher than those for the predicted answers, implying room for improvement in the generated responses of the language models.

\begin{table}[tb]
    \centering
    \caption{Average Human ratings of Psy-LLM responses, only with the two AI-powered versions. \label{table:human-eval-1}}
    \begin{tabular}{@{}ccc@{}}
    \toprule
    Metrics & WenZhong & PanGU  \\ \midrule
    Helpfulness & 3.56  & 3.87\\
    Fluency  & 4.14 & 4.36   \\
    Relevance & 3.87 & 4.09 \\
    Logic & 3.63 & 3.83 \\
    \bottomrule
    \end{tabular}
\end{table}

\begin{table}[tb]
    \centering
    \caption{Average Human ratings of Psy-LLM responses, alongside the ground-truths from datasets. \label{table:human-eval-2}}
    \begin{tabular}{@{}cccc@{}}
    \toprule
    Rating Metrics & WenZhong & PanGU & Ground Truth \\ \midrule
    Helpfulness & 3.45  & 3.54 & 4.52\\
    Fluency  & 3.95 & 4.12 & 4.83   \\
    Relevance & 3.77 & 3.96 & 4.72 \\
    Logic & 3.61 & 3.75 & 4.56 \\
    \bottomrule
    \end{tabular}
\end{table}

\section{Web-Interface for Accessible Online Consultation}\label{sec:web}

One of the primary objectives was to explore the provision of online AI-powered consultation and question-and-answer services in psychology. We adopted a distributed architecture, separating the model's front-end, back-end, and computing servers into modular components. Each module was developed with distinct responsibilities, allowing for easier upgrades and interchangeability of combinations. Communication between the modules was achieved through API interactions, enabling them to function independently without relying on the internal functionality of other modules.

Furthermore, we placed a strong emphasis on security during the design process. We implemented measures to encrypt and protect our modular systems at a product level. The common API interface was productised and encrypted, ensuring secure communication between the components. Additionally, we implemented the HTTPS web system architecture, enhancing security by encrypting each cloud server with TLS (SSL).
By adopting a distributed and modular approach and prioritising security, we aimed to address the challenges of hosting a large-scale online consultation service model. These design choices allowed for flexibility, scalability, and enhanced security in our system architecture, contributing to our project's overall success and reliability.

\subsection{Web Technologies}

We utilised the following services and technologies for our website development:

\begin{itemize}
    \item \emph{ReactJS}: ReactJS was our front-end framework due to its extensive library support. ReactJS offers a wide range of reusable components and follows a modular, component-based architecture, making designing and enhancing the front end easier. ReactJS is responsive and provides excellent cross-platform support.
    \item \emph{AWS Amplify}: AWS Amplify is a rapid front-end deployment service provided by Amazon. It enables us to quickly deploy the front end of our website and seamlessly communicate with other system components. Amplify provides fully managed CI/CD (Continuous Integration/Continuous Deployment) and hosting, ensuring fast, secure, and reliable encryption services.
    \item \emph{Google Domain}: We utilised Google Domain services for secure encapsulation of our EC2 host DNS.
    \item \emph{Amazon EC2}: EC2 provides virtual server instances with highly available underlying designs. It offers reliable, scalable, and flexible access in terms of cost and performance. EC2 provides powerful computing resources and pre-configured environments, making it an excellent choice for running large models. Its robust network performance and high-performance computing clusters allow for high throughput and low-latency online processing. We used simple Flask-based scripts to handle concurrent requests.
    \item \emph{Python \& Flask}: We used Python as our scripting language to run the models and build APIs. Flask, a web framework written in Python, was used for creating API endpoints and handling request-response interactions.
    \item \emph{Apache}: We used Apache, an open-source web server software, for configuring port forwarding, reverse proxies, and listening.
    \item \emph{Let's Encrypt \& Certbot}: We employed Let's Encrypt and Certbot for TLS (Transport Layer Security) encryption, ensuring secure communication between the website and users.
\end{itemize}

\begin{figure}[tb]
    \centering
    \includegraphics[width=1.0\linewidth]{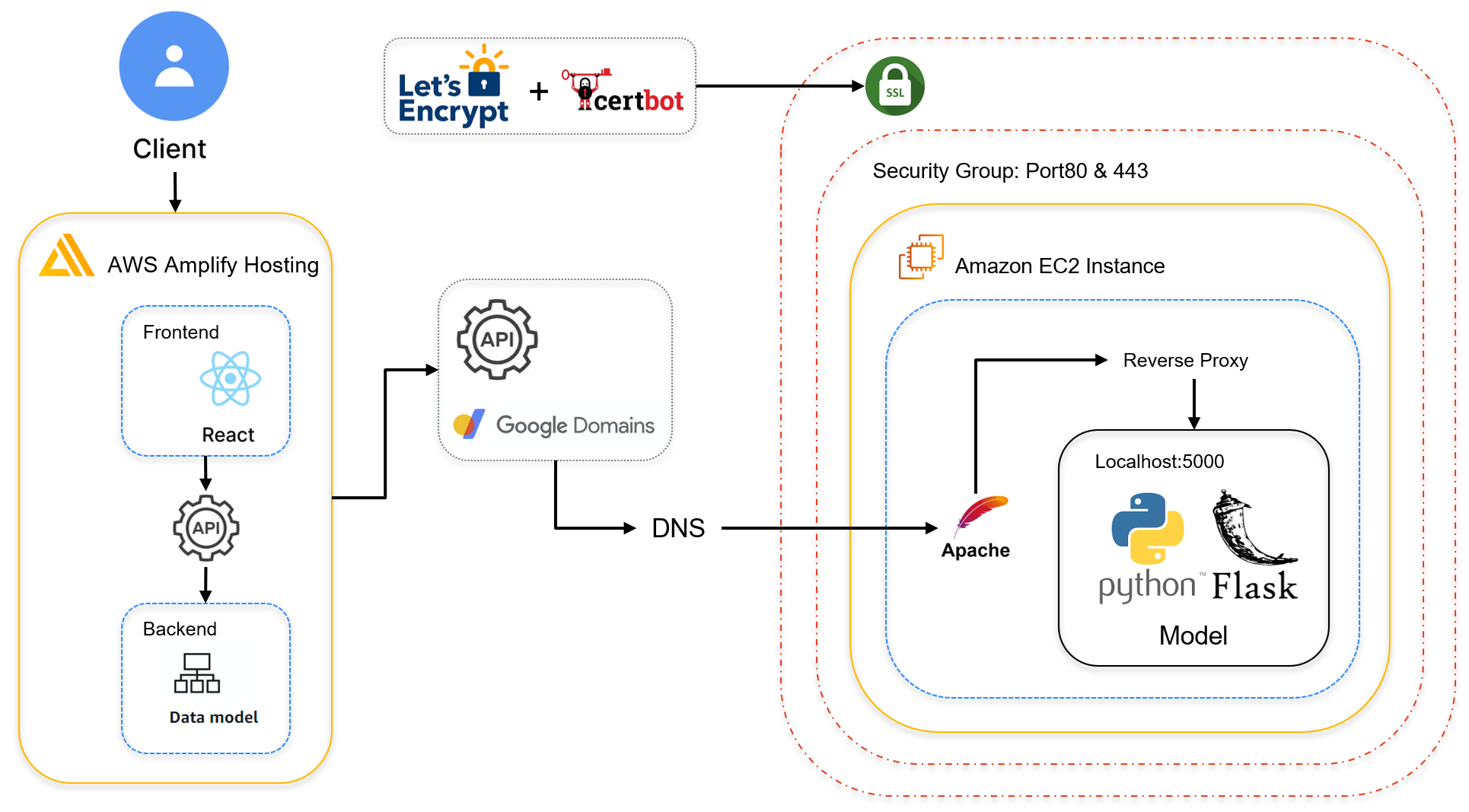}
    \caption{Online Web-frontend Architecture}
    \label{fig:web-arc}
\end{figure}
The diagram in~\cref{fig:web-arc} illustrates the architecture of our independently developed web system for the cloud-based site. The website's user interface is accessible through the front end, deployed on the AWS Amplify service. Built on the ReactJS framework, the front-end communicates with the back-end database through an internal API, enabling storage of user evaluation data for model effectiveness optimisation. The database is hosted within Amplify Hosting. The standalone website interacts with the model runtime server via a public API. To ensure privacy and protect the host address, we register a public domain name through Google Domains and link it to the host server's DNS.

The pre-trained model is deployed on an Amazon EC2 instance host configured as an AWS Linux virtual server. The model uses Python code and Flask scripts, allowing for local server calls. Apache is used for HTTP reverse proxy communication, forwarding external model input data to the local server where the model is waiting and generating results.
To provide secure HTTPS encryption for the web products deployed on AWS Amplify, we employ TSL encryption for the EC2 instance DNS addresses. This is achieved using Let's Encrypt and Certbot as cost-effective alternatives to commercial SSL certificates.

\begin{figure}[tb]
    \centering
    \includegraphics[width=0.5\linewidth]{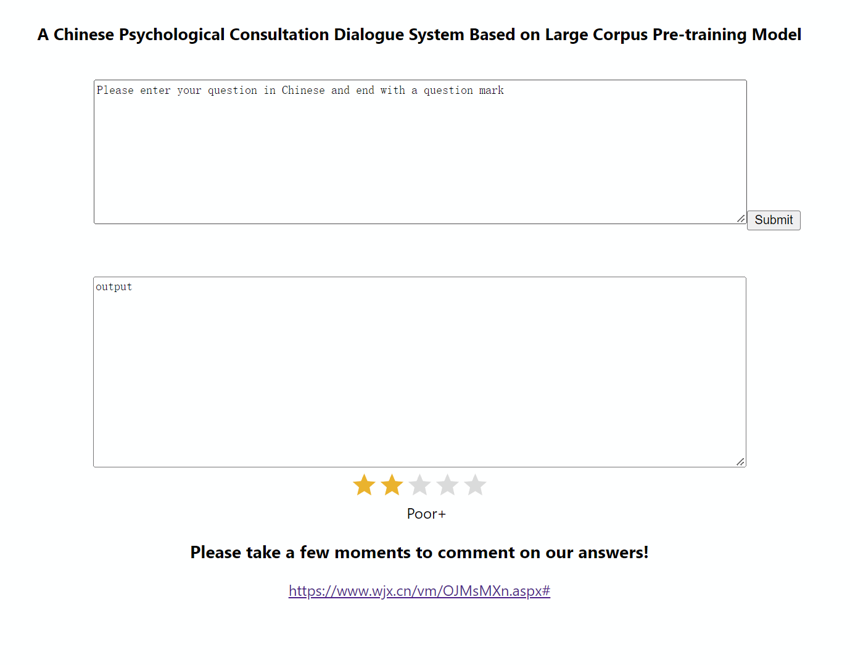}
    \caption{Website Initial Status}
    \label{fig:web-init}
\end{figure}

\begin{figure}[tb]
    \centering
     \begin{subfigure}[t]{0.5\textwidth}
         \centering
        \includegraphics[width=\linewidth]{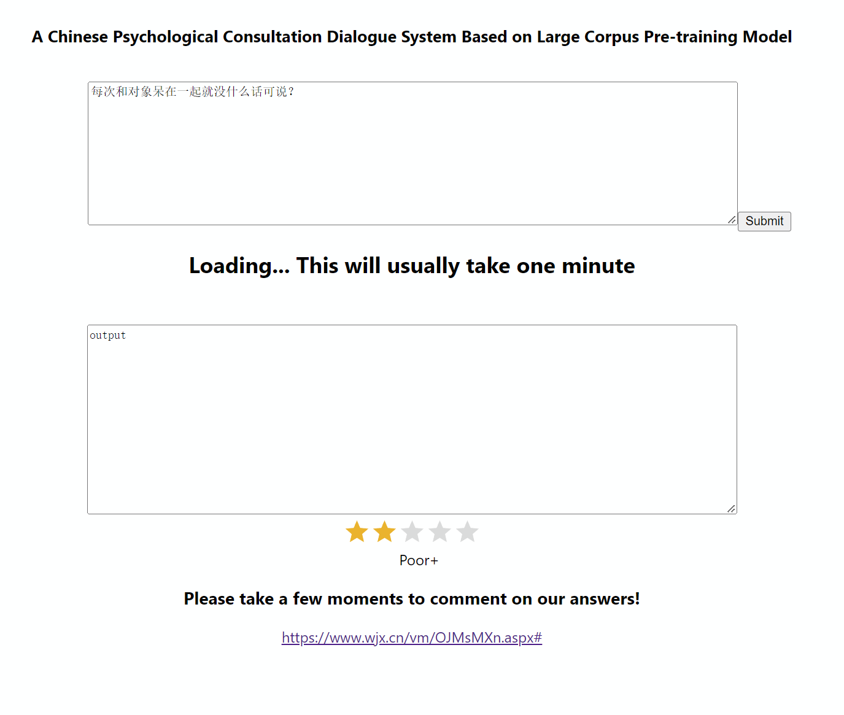}
        \caption{Website Loading Status}
        \label{fig:web-load}
     \end{subfigure}%
     \begin{subfigure}[t]{0.5\textwidth}
         \centering
        \includegraphics[width=\linewidth]{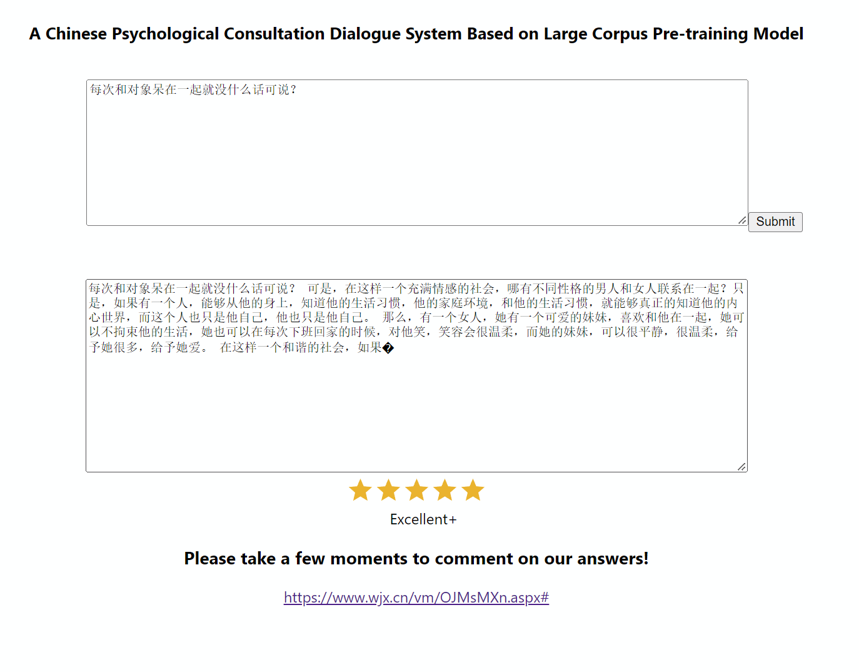}
        \caption{Website Result Status}
        \label{fig:web-done}
     \end{subfigure}
     \caption{Web Front-end for the Online Psychological Consultation}
\end{figure}

The website's front end is designed with simplicity, featuring an input box for users to enter Chinese questions, as depicted in~\cref{fig:web-init}. Upon submission, the system communicates with the back-end model through the API. It awaits the completion of model processing (\cref{fig:web-load}) before returning the results to the output box, as depicted in (\cref{fig:web-done}). Users can rate the results using the built-in rating system, and there is a link to an additional evaluation site at the bottom of the page.

%% file: src/8.discussion.tex
\section{Discussion}

The discussion section provides a comprehensive analysis of the project outcome, product perspective, website perspective, model perspective, and evaluation perspective.

\subsection{Project Outcome}

We have successfully developed and implemented an effective chatbot for mental health counselling. Through the training and fine-tuning large-scale Chinese pre-training models on mental health datasets, the chatbot has acquired valuable knowledge in psychology, enhancing its ability to provide counselling services. The deployment of the chatbot on a website interface has created a convenient and accessible platform for users seeking mental health support. Although the chatbot is currently in its prototype stage, our project demonstrates the feasibility of building an AI-based counselling system. It is a valuable reference for future research and development in this area.

From a model perspective, our evaluation results demonstrate the superiority of the \emph{PanGu} model over the \emph{WenZhong} model, as expected due to its larger size and advanced architecture. The \emph{PanGu} model's design contributes to its outperformance, particularly its incremental learning ability and enhanced natural language understanding capabilities. However, both models fall short of achieving human-level performance, which can be attributed to the quality of the training dataset and the inherent limitations of autoregressive language models. Enhancing the dataset quality and exploring alternative language model architectures hold promise for addressing these limitations and further improving performance.

\subsection{Evaluation Perspective}

The human evaluation results indicate that both the \emph{PanGu} model and the \emph{WenZhong} model have yet to reach human-level performance. Even though training the models on our dataset crawled from websites, the predicted answers strongly focus on psychological content but lack logical coherence. One potential reason for this is the quality of our dataset, which may need to be higher to provide comprehensive and reliable training examples. Although we conducted human evaluation during the data cleaning stage, the sheer volume of data made it challenging to cover every instance. To address this, we recommend thoroughly evaluating the website data before crawling to ensure a higher-quality dataset.

Another factor impacting human evaluation is the limited computing conditions during model training. Our model requires a specific training environment and numerous parameters, making it time-consuming to adjust and fine-tune it effectively. We could not optimise the parameters and achieve optimal testing results with our current resources. Consequently, the model's performance may have been hindered by these limitations.
Furthermore, the autoregressive nature of both the \emph{PanGu} model and the \emph{WenZhong} model poses challenges in comprehending contextual information. As autoregressive language models, their training processes are unidirectional, focused on modelling the joint probability from left to right. The next predicted word is solely based on the preceding word, limiting their ability to capture information from broader contexts. This lack of contextual background reference makes it difficult for language models to handle reading comprehension tasks like humans.

In summary, the evaluation results shed light on the areas where improvements can be made. Enhancing the dataset quality through pre-evaluation and addressing the limitations of our computing conditions are crucial steps toward advancing the model's performance. Additionally, exploring alternative language model architectures that can effectively capture contextual information may contribute to bridging the gap between model-generated responses and human-level performance.

\subsection{Product and Practicality Perspective}

The performance of the online consultation service indicates its significant potential for streamlining mental health support with minimal resources. The user experience has been a priority in product design, and the cloud infrastructure deployment ensures easy access via mobile devices. As part of our future improvements, we plan to incorporate an automatic emotion recognition system into the website, enabling the identification of users in distress and facilitating timely intervention. The design and development of our product hold substantial societal value in the mental health support field, providing a promising avenue for further exploration and refinement.

Regarding the website, we have designed and implemented a modern, cloud-based network architecture that boasts lightweight, scalable, and highly secure features. This architecture allows for low-cost, large-scale model computing sites, enabling widespread accessibility to AI-based question-and-answer services. Our approach serves as a reference for small organisations and enterprises with limited resources, showcasing the possibilities of deploying AI capabilities effectively.

%% file: src/9.limitations.tex
\section{Limitations and Future Works}

While we have presented promising results with our Psy-LLM model for usage in assisting mental health workers, our study is exploratory in nature, and hence, there exist numerous limitations that we would like to raise in the following.

\subsection{Model Capability and Usage in Real-World}
While there are numerous benefits in deploying an AI-powered Large Language Model for supporting the demand in the mental health sector, one should consider several ethical and practical issues.
Firstly, as a language-based model, the model's output is based purely on the input text. 
However, studies have shown that nonverbal communication is one of the key factors in counselling outcome~\citep{hill1981nonverbal}.
In fact, a well-trained counsellor can often pick up subtle cues even when there is a lack of response from the patient.
Standalone LLM models like Psy-LLM cannot address such an issue (unless techniques like facial emotion detection from the computer vision community are integrated as a unified system~\citep{jaiswal2020facial}).
Furthermore, rapport-building with clients is often a crucial step in clinical psychology.
However, an AI-based model would face severe difficulties in building trusted client relationships.
As a result, it is critical to realise that such an AI-powered system cannot replace real-world counselling setups.
A practical approach would be to pair the model output under the supervision of a trained counsellor as a good psychoeducational tool.
The model output can be used as an initial guideline or suggestion for assisting human counsellors in providing useful and trusted consultations with patients.

\subsection{Data Collections}

Several strategies can be implemented in future work to overcome the limitations in data collection. Firstly, to address the issue of anti-crawler rules on different websites, developing a more robust and adaptable crawler that can handle different anti-crawler mechanisms would be beneficial. The access limitation could involve implementing dynamic IP rotation or utilising proxies to avoid IP blocking. Machine learning techniques, such as automatic rule extraction or rule adaptation, could also help automate handling anti-crawler mechanisms.

Incorporating more advanced data-cleaning techniques can also improve the quality of the crawled data. More advanced data-cleaning procedures may involve NLP methods, such as entity recognition, part-of-speech tagging, and named entity recognition to identify and filter out irrelevant or noisy data. Additionally, leveraging machine learning algorithms, such as anomaly detection or outlier detection, can aid in identifying and removing low-quality or erroneous data points.
In terms of dataset standardisation, establishing a unified standard for data generation in the online domain would greatly facilitate the cleaning process. This could involve collaborating with website administrators or data providers to develop guidelines or formats for data representation. Furthermore, using human annotators or experts in the domain to manually review and clean a subset of the dataset can provide valuable insights and ensure a higher-quality dataset.

However, it is important to acknowledge that achieving a completely clean dataset is challenging, particularly when dealing with large-scale datasets. As such, future work should strike a balance between the manual review and automated cleaning techniques while also considering the cost and scalability of the data cleaning process.

\subsection{Model Improvement}

Increasing the scale of model training by utilising larger models or ensembles of models can enhance the performance and capabilities of the chatbot. Larger models can capture more nuanced patterns and relationships in the data, leading to more accurate and coherent responses.
Exploring different model architectures beyond autoregressive language models may provide valuable insights. Bidirectional models (e.g. Transformer-XL) or models that incorporate external knowledge sources (e.g. knowledge graphs) can improve the chatbot's contextual understanding and generate more informative responses.
Moreover, integrating feedback mechanisms into the training process can help iteratively improve the chatbot's performance. This could involve collecting user feedback on the generated responses and incorporating it into the model training through reinforcement learning or active learning.

Several disadvantages were also identified in the LLM architecture. Firstly, the maximum likelihood training approach of the \emph{WenZhong} model is susceptible to exposure bias, which occurs when samples are drawn from the target language distribution. This bias can lead to errors for which researchers have yet to find effective solutions. Additionally, training the \emph{WenZhong} model multiple times can significantly decrease its quality.
Furthermore, the \emph{WenZhong} model follows an autoregressive architecture, which models joint probability from left to right. This unidirectional training process limits its ability to capture information from all contexts, particularly hindering its performance in tasks requiring reading comprehension that rely on contextual background references.
Similar to the \emph{WenZhong} model, the \emph{PanGu} model also exhibits autoregressive characteristics. Although it inherits the ability to estimate the joint probability of language models, it suffers from the same limitations of unidirectional modelling. It lacks bidirectional context information and may produce duplicate results requiring resolve deduplication.

We also have reservations about the Jieba tokeniser used in the \emph{PanGu} model. Its performance and tokenisation ability need to handle complex Chinese tokenisation accurately. Furthermore, as neural networks and pre-trained models advance, Chinese NLP tasks increasingly demonstrate that tokenisation is only sometimes necessary. Large models can effectively learn character-to-character relationships without word segmentation. For instance, Google is considering discarding tokenisation and using bytes directly. Adopting a more flexible tokeniser could make the model more suitable for various industrial applications, even when sacrificing some performance.

\subsection{User Experience and User Interface}
Enhancing the chatbot's user experience and user interface can significantly impact its adoption and effectiveness. Future work should focus on improving the simplicity, intuitiveness, and accessibility of the website interface. This includes optimising response times, refining the layout and design, and incorporating user-friendly features such as autocomplete suggestions or natural language understanding capabilities.

Furthermore, personalised recommendations and suggestions to users based on their preferences and previous interactions can enhance the user experience. Techniques like collaborative filtering or user profiling can enable the chatbot to understand better and cater to individual user needs.
Usability testing and user feedback collection should be conducted regularly to gather insights on user preferences, pain points, and suggestions for improvement. Iterative design and development based on user-centred principles can ensure that the chatbot meets user expectations and effectively addresses their mental health support needs.

\subsection{Ethical Considerations and User Privacy}
As with any AI-based system, ethical considerations and user privacy are paramount. Future work should address these concerns by implementing robust privacy protection mechanisms and ensuring transparency in data usage. This includes obtaining explicit user consent for data collection and usage, anonymising sensitive user information, and implementing strict data access controls.
Developing mechanisms to handle potentially sensitive or harmful user queries is crucial. The chatbot should have appropriate safeguards and guidelines to avoid providing inaccurate or harmful advice. Integrating a reporting system where users can report problematic responses or seek human intervention can help mitigate potential risks.
Furthermore, monitoring and auditing the chatbot's performance and behaviour can help identify and rectify biases or discriminatory patterns. Regular evaluations by domain experts and user feedback analysis can improve the chatbot's reliability, fairness, and inclusivity.

While this project has made significant progress in developing an AI-based chatbot for mental health support, there are various limitations and areas for improvement. Overcoming challenges related to data quality, model performance, ethical considerations, and user experience will contribute to the overall effectiveness and reliability of the chatbot. By addressing these limitations and exploring future research directions, we can continue to advance the field of AI-powered mental health support systems and provide valuable assistance to individuals in need.

\section{Conclusion}

In conclusion, our project on Psy-LLM, an exploratory study on using Large Language Models as an assistive mental health tool, has been successfully completed and implemented. While there are areas identified for improvement based on specific evaluation indicators, we are confident that with improved equipment conditions, we can enhance the performance of this platform. The experimental results obtained from this project hold significant potential to contribute to the fields of supportive natural language generation and psychology, driving advancements at the intersection of these domains. The deployment of such a system offers a practical approach to promoting the overall mental well-being of our society by providing timely responses and support to those in need.